\documentclass[lettersize,journal]{IEEEtran}
\usepackage{amsmath,amsfonts}
\usepackage{algorithmic}
\usepackage{algorithm}
\usepackage{array}
\usepackage[caption=false,font=normalsize,labelfont=sf,textfont=sf]{subfig}
\usepackage{textcomp}
\usepackage{stfloats}
\usepackage{url}
\usepackage{verbatim}
\usepackage{graphicx}
\usepackage{cite}
\usepackage{pifont}
\usepackage{bbm}
\usepackage{amsmath}
\usepackage{amssymb}
\usepackage{booktabs}
\usepackage{multirow}
\usepackage{makecell}
\usepackage{xcolor}
\usepackage{url}
\usepackage{caption}
\usepackage{subcaption}

\usepackage{colortbl}  %彩色表格需要加载的宏包
\definecolor{mygray}{rgb}{1,0.95,0.97}

\usepackage[capitalize]{cleveref}
\crefname{section}{Sec.}{Secs.}
\crefname{table}{Tab.}{Tabs.}

\begin{document}

\title{From Mapping to Composing: A Two-Stage Framework for Zero-shot Composed Image Retrieval}

\author{Yabing Wang, Zhuotao Tian, Qingpei Guo, Zheng Qin, Sanping Zhou, Ming Yang, Member, IEEE, Le Wang, Senior
Member, IEEE
% \author{IEEE Publication Technology,~\IEEEmembership{Staff,~IEEE,}
        % <-this % stops a space
\thanks{Y. Wang, Z. Qin, S. Zhou, and L. Wang are with the  Institute of Artificial Intelligence and Robotics, Xi'an Jiaotong University, Xi'an, Shaanxi 710049, China. (Email: wyb7wyb7@gmail.com)}
\thanks{Z. Tian is with the Harbin Institute of Technology, Shenzhen, 518071, China.}
\thanks{Q. Guo and M. Yang are with Ant Group, Hangzhou, 310000, China.}
% \thanks{This paper was produced by the IEEE Publication Technology Group. They are in Piscataway, NJ.}% <-this % stops a space
\thanks{Manuscript received X, 2025; revised X, 2025.}}

% The paper headers
\markboth{Journal of \LaTeX\ Class Files,~Vol.~X, No.~X, X~2025}%
{Shell \MakeLowercase{\textit{et al.}}: A Sample Article Using IEEEtran.cls for IEEE Journals}

\IEEEpubid{0000--0000/00\$00.00~\copyright~2025 IEEE}
% Remember, if you use this you must call \IEEEpubidadjcol in the second
% column for its text to clear the IEEEpubid mark.

\maketitle

\begin{abstract}
Composed Image Retrieval (CIR) is a challenging multimodal task that retrieves a target image based on a reference image and accompanying modification text. Due to the high cost of annotating CIR triplet datasets, zero-shot (ZS) CIR has gained traction as a promising alternative. Existing studies mainly focus on projection-based methods, which map an image to a single pseudo-word token. However, these methods face three critical challenges: (1) insufficient pseudo-word token representation capacity, (2) discrepancies between training and inference phases, and (3) reliance on large-scale synthetic data. To address these issues, we propose a two-stage framework where the training is accomplished from mapping to composing. In the first stage, we enhance image-to-pseudo-word token learning by introducing a visual semantic injection module and a soft text alignment objective, enabling the token to capture richer and fine-grained image information. In the second stage, we optimize the text encoder using a small amount of synthetic triplet data, enabling it to effectively extract compositional semantics by combining pseudo-word tokens with modification text for accurate target image retrieval.
The strong visual-to-pseudo mapping established in the first stage provides a solid foundation for the second stage, making our approach compatible with both high- and low-quality synthetic data, and capable of achieving significant performance gains with only a small amount of synthetic data.
Extensive experiments were conducted on three public datasets, achieving superior performance compared to existing approaches.

\end{abstract}


\section{Introduction}

\IEEEPARstart{U}{nlike} traditional cross-modal retrieval \cite{luo2025cross,wang2024dual,dong2022reading,wang2024semantics,ma2024cross,liu2023efficient}, composed image retrieval (CIR) offers greater flexibility by allowing users to specify a reference image and retrieve the target image based on a modification description.
Despite the impressive performance of existing CIR methods \cite{wen2023target,huang2024dynamic,ventura2024covr}, their reliance on costly manually annotated triplets (reference image, modified text, and target image) restricts the availability of large-scale training data, thereby limiting their scalability and generalization capabilities in real-world applications.

\begin{figure}[tb!]
\centering\includegraphics[width=0.9\columnwidth]{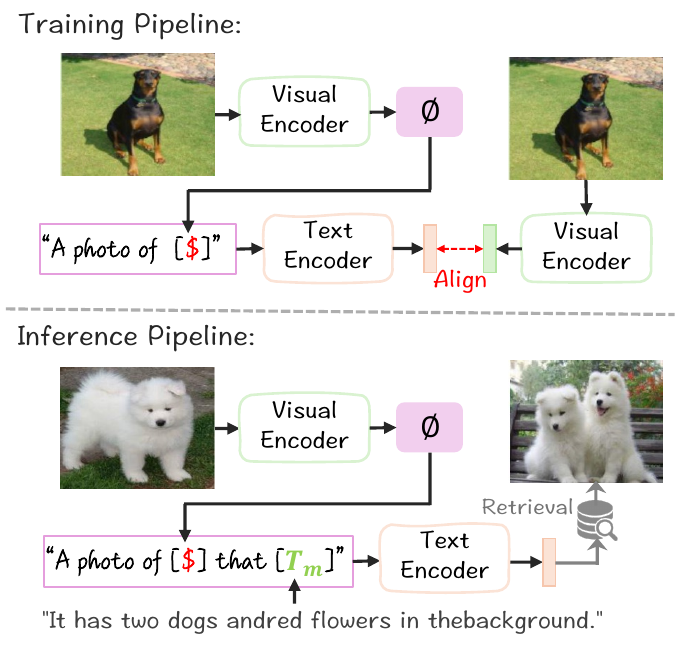}
\vspace{-4mm}
\caption{
Training and inference pipelines of the projection-based method.  During training, the model focuses on mapping images to pseudo-word tokens, while during inference, the model needs to combine the pseudo-word token with real words to generate the composed query. ``$\phi$" refers the mapping network, ``\$" indicates the pseudo-word token, and ``$T_m$" denotes the modified text. 
}\label{fig:title-pic}
\vspace{-2mm}
\end{figure}

To overcome this problem, zero-shot (ZS) CIR has emerged as a promising alternative that eliminates the need for annotated training triplets. 
Considering the easy availability of large-scale image-text data 
e.g., CC3M~\cite{sharma2018conceptual}
,current approaches~\cite{saito2023pic2word,tang2024context,gu2024language,jiang2024hycir} typically reformulate the CIR task as a text-to-image retrieval problem by learning a mapping network $\phi$. As shown in Figure~\ref{fig:title-pic}, these methods aim to convert an image into a pseudo-word token \$, leveraging the multimodal alignment capabilities of vision-language models (VLMs), such as CLIP \cite{radford2021learning}. During inference, the pseudo-word token is combined with the modification text $T_m$ to produce a composed query, e.g., “a photo of \$ that $T_m$”,  which is then used to retrieve the target image.
Despite making significant progress, these methods encounter the following challenges:

\begin{figure}[tb!]
\centering\includegraphics[width=1\columnwidth]{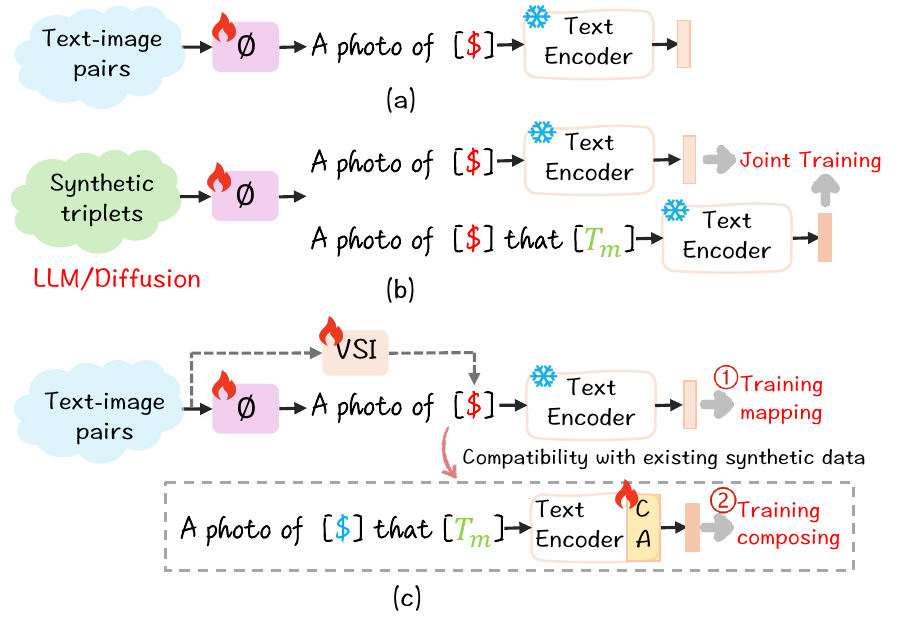}
\vspace{-6mm}
\caption{The paradigms of the current projection-based method:  (a) The baseline approach, which primarily focuses on mapping images to pseudo-word tokens. (b) Method leveraging LLMs or diffusion models to generate large-scale synthetic data, training mapping, and composing simultaneously. (c) We propose a two-stage framework that decouples the learning process into mapping and composing stages.
It enhances the model’s compositional understanding capability using only a small amount of synthetic data.  ``TL" represents the designed token learner module, ``VSI'' and ``CA" denote the visual semantic injection module and composed adapter, respectively. 
}\label{fig:title-pic-2}
\end{figure}

\IEEEpubidadjcol
(1) \textit{Inadequate Representation:} 
An image often conveys more information than a single word can capture. As illustrated in Figure~\ref{fig:title-pic-2}(a), these methods that rely on global visual features to generate pseudo-word tokens may struggle to fully represent the complex and fine-grained content of an image.
Although some approaches \cite{tang2024context, du2024image2sentence} design token learners to address this limitation, the inherent modality gap can still distort the original and discriminative semantics of images during the projection process, as highlighted in \cite{jang2025spherical}.

(2) \textit{Training \& Inference Discrepancy:} 
Throughout the training phase, these methods focus on mapping an image to a word token, while during inference, the task shifts to generating a composed query by combining the word token with the modification text.
Nevertheless, VLMs struggle to effectively encode the composed query, as they face challenges in integrating the pseudo-word token with the modification text.

(3) \textit{Large-scale Synthetic Data:} 
Some methods \cite{gu2024compodiff, lin2024fine, byun2024reducing,jiang2024hycir}, as shown in Figure~\ref{fig:title-pic-2}(b), leverage LLMs or diffusion models to generate synthetic data in order to address the above discrepancy.
However, due to the inherent noisiness of synthetic data (e.g., caused by the hallucination problem in LLMs), these methods often require large-scale datasets (e.g., 18.8M data generated in \cite{gu2024compodiff}) for training.
Yet, generating such large-scale synthetic data is both time-consuming and resource-intensive, posing significant challenges to the scalability of these approaches in practical scenarios.

To address the challenges mentioned above, we propose a two-stage framework, called TSCIR, where training is divided into a mapping learning stage followed by a compositional learning stage, as shown in Figure~\ref{fig:title-pic-2}(c). 
In the first stage, we leverage large-scale, easily accessible multi-modal data to train the model to map images to pseudo-word tokens, enabling it to focus solely on capturing rich and discriminative image semantics.
In the second stage, we use existing synthetic data to train the model to compose the pseudo-word token with the modification text, thereby learning compositional semantics.
This staged design allows the model to first acquire a comprehensive understanding of visual semantics, and then build upon it to enhance its compositional learning capabilities. \textit{The underlying motivation is that, in CIR, not all visual semantics need to be used as reference conditions; however, 
only once the pseudo-word token has captured the full visual information can the model then adaptively select the necessary semantics based on the modified text.
}

Furthermore, by decoupling the two learning objectives, our framework offers greater flexibility and generalizability. The first stage lays a solid foundation by learning a strong visual-to-pseudo mapping, which facilitates the second stage in effectively learning compositional semantics. As a result, our method is compatible with both high- and low-quality synthetic data, and achieves strong performance using only a small amount of synthetic data, for example, just $0.5\%$ randomly sampled from SynTriplets18M~\cite{gu2024compodiff}.

Specifically, in the first stage, we design the visual semantic injection module into the text encoder to capture more comprehensive visual information, and employ a soft text alignment objective to align pseudo-word tokens with real words. This enables the model to learn a more strong mapping between the image and the pseudo-word token. In the second stage, we introduce several composing adapters along with a hard learning strategy to encourage the model to learn compositional semantics. Notably, these adapters contain only $\sim1\text{M}$ parameters, introducing minimal computational overhead.
Extensive experiments on three popular benchmarks (i.e., Fashion-IQ \cite{wu2021fashion}, CIRR \cite{liu2021image}, and CIRCO \cite{baldrati2023zero}), achieving state-of-the-art results across all evaluated datasets. Even without the second-stage training, our method already achieves comparable performance to approaches relying on large-scale synthetic data, demonstrating the strong pseudo-word token representation in the first stage.
Our contributions can be summarized as follows:

(1) We propose a two-stage framework, TSCIR, consisting of a mapping learning stage followed by a compositional learning stage. This design enables the model to learn a more comprehensive image-to-pseudo-word mapping and enhances its capability for compositional understanding.

(2) The two-stage design allows our method to flexibly incorporate various existing synthetic datasets generated by previous works. It achieves strong performance with only a small amount of synthetic data, and still demonstrates superior performance even without any synthetic data.

(3) We conduct extensive experiments on three ZS-CIR benchmarks, Fashion-IQ\cite{wu2021fashion}, CIRR\cite{liu2021image}, and CIRCO~\cite{baldrati2023zero}, demonstrating the effectiveness and generalization of our proposed method.

\section{Related Works}

\textbf{Composed Image Retrieval.}
Composed Image Retrieval (CIR) aims to retrieve a target image based on a reference image and a modification description. Compared to traditional cross-modal retrieval \cite{wang2022cross,wang2024cl2cm,wang2024multimodal,wang2024semantics,cai2024cross,jin2025revealing,pu2025deep,wu2025large}, CIR offers greater flexibility by allowing users to specify fine-grained changes to the reference image via natural language.
Existing methods \cite{yang2024sda,feng2024improving,zhao2022progressive,huang2024dynamic,baldrati2022conditioned,ventura2024covr,wen2023target} primarily focus on fusing the representations of the modification description and the reference image, while aligning the fused representations with the target image representations.
For example, TIRG \cite{Vo_Jiang_Sun_Murphy_Li_Fei-Fei_Hays_2019} and PL4CIR \cite{zhao2022progressive} introduce gates or weight mechanisms to dynamically determine the importance of image and text within the hybrid-modality query for improved retrieval performance. Similarly, CaLa \cite{jiang2024cala} designs two complementary association modules to capture the underlying relationships within triplet data.
While these methods have achieved significant progress, their reliance on costly manually annotated triplet datasets limits their scalability and broader applicability.

\textbf{Zero-Shot Composed Image Retrieval.} To address the aforementioned limitation, Zero-Shot Composed Image Retrieval (ZS-CIR) methods \cite{gu2024compodiff,gu2024language,yang2024ldre,jang2025spherical,sun2023training,yang2024semantic,zhang2024zero} have emerged as a promising alternative. 
The mainstream solution, namely the projection-based method, addresses the CIR task by converting it into a standard text-to-image retrieval problem through the pre-training of a textual inversion network.
For example, Pic2word~\cite{saito2023pic2word} is a pioneering approach that learns a mapping layer to project image features into the text embedding space. 
Subsequent works have sought to enhance textual inversion training, such as SEARLE~\cite{baldrati2023zero}, integrate pseudo-word tokens with GPT-based captions for data generation, while Context-I2W~\cite{tang2024context} enhances the mapping by incorporating context-dependent networks to adaptively transform image information into pseudo-words.
However, these methods struggle with the discrepancy between training and inference phases, making it difficult to integrate textual and visual features effectively for query construction. To mitigate this, some methods \cite{gu2024compodiff,jiang2024hycir,liu2023zero} leverage large language models or stable diffusion to generate pseudo-triplet datasets, enriching the training data. Yet, these methods rely on large-scale synthetic data and may result in suboptimal pseudo-word tokens.
In this paper, we propose a two-stage training framework where the training is accomplished from mapping to composing. Our approach enables the pseudo-word token to comprehensively capture image content and effectively generate composed queries by combining the pseudo-word token with modified text.

\section{Method}

\begin{figure*}[t!]
\centering\includegraphics[width=2\columnwidth]{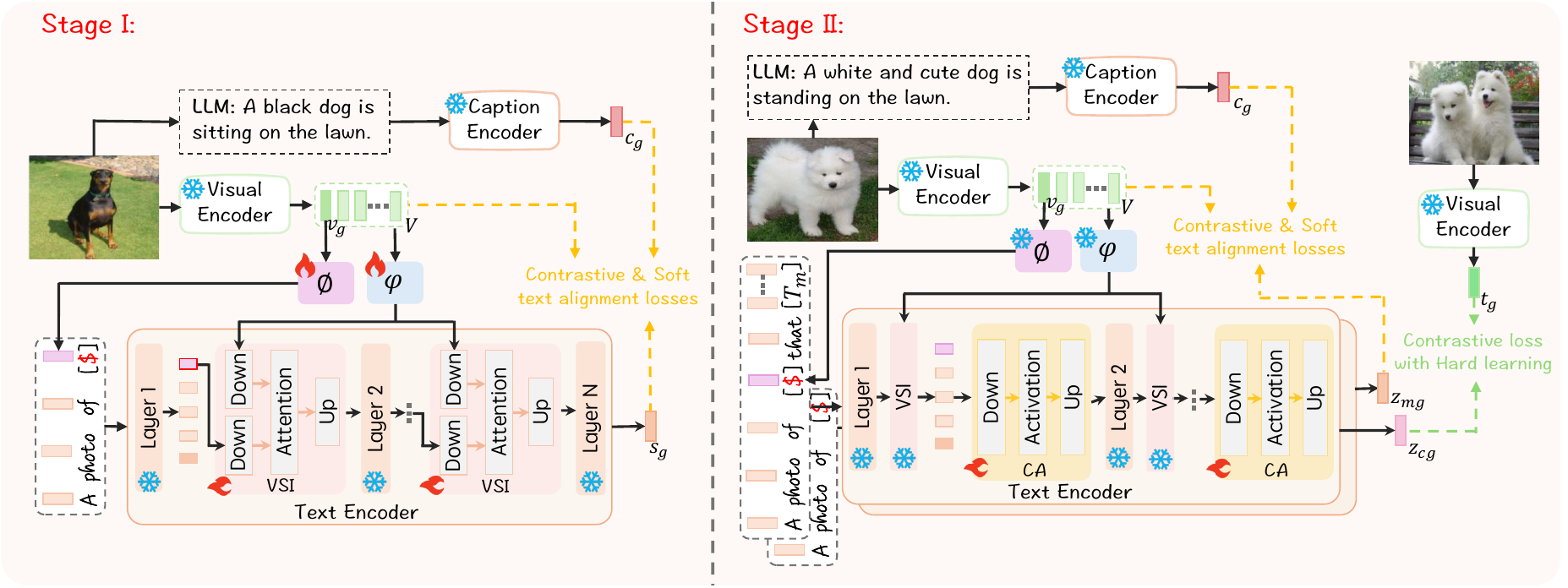}
\vspace{-3mm}
\caption{The framework of our proposed method comprises two stages: mapping learning (left) and composing learning (right). 
In the first stage, to comprehensively capture visual information, we introduce the visual semantic inject module (VSI), which can be integrated into various layers of the text encoder to continuously inject visual semantics. Additionally, a soft text alignment loss is applied to ensure the pseudo-word token aligns well with real words.
In the second stage, we incorporate several composing adapters (CA) and adopt a hard negative mining strategy to optimize the text encoder, enabling it to effectively encode the composed query by combining the pseudo-word token with the modification text.
}\label{fig:framework-pic}
% \vspace{-4mm}
\end{figure*}

In this section, we provide a detailed description of the proposed two-stage framework, as illustrated in Figure~\ref{fig:framework-pic}. We first introduce the visual semantic mapping stage (Section~\ref{sec:mapping}), which aims to map an image to a pseudo-word token that captures rich and comprehensive semantic information.
Subsequently, in Section~\ref{sec:composed}, we describe the compositional learning stage, which focuses on effectively encoding the composed query by integrating the pseudo-word token with the modification text.

\subsection{Visual Semantic Mapping}
\label{sec:mapping}

In this stage, we aim to learn a pseudo-word token that encapsulates the image semantics more comprehensively by mapping the image into the textual embedding space.
Specifically, given an image set $X = {\{x\}}^\mathcal{B}_{i=1}$, where $\mathcal{B}$ denotes mini-batch, we first use the image encoder $\mathcal{F}_{img}$ to extract the image representations.  These features are then input into mapping network~$\phi(\cdot)$ to obtain a pseudo-word token, denoted as $\mathbf{w} \in \mathbb{R}^d$:
\begin{gather}
    \mathbf{v_g}, \mathbf{V} = \mathcal{F}_{img}(X) \\
    \mathbf{w} = \phi(\mathbf{v_g})
\end{gather}
where $\mathbf{v_g} \in \mathbb{R}^d$ is the [CLS] token, representing the global image features, and $\mathbf{V} \in \mathbb{R}^{m \times d}$ represents the patch features with $m$ indicating the number of patches.
Subsequently, we employ the template $P_1$ \textit{“A photo of \$”} for tokenization, replacing the placeholder \$ with pseudo-word token $\mathbf{w}$. This process results in a sequence of token embeddings $\mathbf{E} \in \mathbb{R}^{n \times d}$, where $n$ denotes the number of tokens. 
Finally, these embeddings are fed into the text encoder $\mathcal{F}_{txt}$ to extract the text representations~$\mathbf{S} \in \mathbb{R}^{n \times d}$:
\begin{gather}
    \mathbf{S} = \mathcal{F}_{txt}(\mathbf{E}) 
\end{gather}

However, relying solely on global image features may fail to capture the full semantic richness of an image. To address this limitation, we introduce the Visual Semantic Injection (VSI) module, which continuously injects visual semantics into the text encoder. This design enables the model to incorporate visual information more comprehensively and effectively throughout the encoding process.

\textbf{Visual Semantic Injection.}
We first enhance the local contextual information in the image patch features using a convolutional layer, and then project them into the text embedding space through a mapping network $\varphi(\cdot)$:
\begin{gather}
    \bar{\mathbf{V}} = \varphi(\texttt{Conv}(\mathbf{V}))
\end{gather}

Subsequently, a cross-attention module is employed to extract visual semantics from the projected features and inject them into the pseudo-word tokens $\mathbf{s}_w$, enabling the model to incorporate richer visual information during text encoding.
Taking the $i$-th layer of the text encoder as an example, this process is formulated as:
\begin{gather}
    \mathbf{H}_v^i = \psi_{v}^{i}(\bar{\mathbf{V}}), \
    \mathbf{h}_w^i = \psi_{w}^{i}(\mathbf{s}_w^i), \ 
    \mathbf{h}_g^i = \psi_{g}^{i}(\mathbf{s}_g^i), \\
    \mathbf{c}^i = \texttt{Attention}(\mathbf{h}_w^i+\mathbf{h}_g^i, \mathbf{H}_v^i, \mathbf{H}_v^i) \\
    \hat{\mathbf{s}}_w^i = \psi_{u}^{i}(\mathbf{c}^i) + \mathbf{s}_w^i
\end{gather}
where $\psi^{i}_{v,w,g}(\cdot)$ denotes a multilayer perceptron (MLP) used to project features into a lower-dimensional latent space, while $\psi^{i}_{u}(\cdot)$  is employed to restore the features back to their original dimensionality.
Moreover, $\mathbf{s}_g^i$ represents the [CLS] token, which serves as global semantic guidance and is incorporated into the cross-attention mechanism to enhance the effectiveness of visual semantic injection.

Empirically, we find that integrating visual semantics solely into the pseudo-word tokens led to superior performance, compared to integrating them into the [CLS] token or all word tokens (see Table~\ref{tab:va-position}). This approach ensures that the reference image semantics do not overwhelm the overall composed semantics.

\textbf{Optimization.} 
To promote the final text features to capture the corresponding visual content, we employ contrastive loss to encourage alignment between the text features $\mathbf{s_g}$ and the corresponding image features $\mathbf{v_g}$, defined as:
\begin{gather}
    \mathcal{L}_{map} = \mathcal{L}_{contra}(\mathbf{s_g}, \mathbf{v_g}) + \mathcal{L}_{contra}(\mathbf{v_g},\mathbf{s_g})
\end{gather}

The contrastive loss $\mathcal{L}_{contra}(\mathbf{u}, \mathbf{o})$ is formulated as:

\begin{gather}
\mathcal{L}_{contra}(\boldsymbol{u}, \boldsymbol{o})=-\frac{1}{|\mathcal{B}|} \sum_{i \in \mathcal{B}} \log \frac{\exp \left(\tau \boldsymbol{u}_i^T \boldsymbol{o}_i\right)}{\sum_{j \in \mathcal{B}} \exp \left(\tau \boldsymbol{u}_i^T \boldsymbol{o}_j\right)}
\end{gather}
where $\tau$ is the temperature coefficient.

\textbf{Soft Text Alignment Loss.}
Furthermore, to ensure that the learned text features are well-aligned in the text space and can interact effectively with real words, we introduce a text alignment objective to align the learned features $\mathbf{s_g}$ with the caption features $\mathbf{c_g}$ extracted by the original text encoder.

Given that captions often describe only partial aspects of the visual content, we adopt soft targets to supervise the alignment process. These targets are derived from the similarity distribution between image and caption features, which reflects their semantic correspondence. This strategy mitigates the risk of semantic collapse and helps preserve the richness of the learned representations. The process can be formulated as follows:
\begin{gather}
    P = -\frac{1}{|\mathcal{B}|} \sum_{i \in \mathcal{B}} \frac{\exp \left(\tau \boldsymbol{\mathbf{v_g}}_i^T \boldsymbol{\mathbf{c_g}}_i\right)}{\sum_{j \in \mathcal{B}} \exp \left(\tau \boldsymbol{\mathbf{v_g}}_i^T \boldsymbol{\mathbf{c_g}}_j\right)}, \\
    Q = -\frac{1}{|\mathcal{B}|} \sum_{i \in \mathcal{B}} \frac{\exp \left(\tau \boldsymbol{\mathbf{s_g}}_i^T \boldsymbol{\mathbf{c_g}}_i\right)}{\sum_{j \in \mathcal{B}} \exp \left(\tau \boldsymbol{\mathbf{s_g}}_i^T \boldsymbol{\mathbf{c_g}}_j\right)}, \\
    \mathcal{L}_{sta} = \frac{1}{\mathcal{B}} \sum \limits^\mathcal{B}_{i=1} {KL}(P^i \ || \ Q^i)
\end{gather}
where $KL(||)$ means the KL divergence. Finally, the objective for the first stage is formulated as:

\begin{gather}
\label{eq:stage2_loss}
    \mathcal{L}_{\texttt{stage-I}} = \mathcal{L}_{map} + \alpha \mathcal{L}_{sta}
\end{gather}
where $\alpha$ is a hyperparameter to control the weight of soft text alignment loss.

\subsection{Compositional Learning}
\label{sec:composed}
Through the aforementioned training process, the model has effectively learned the image-to-pseudo mapping. Subsequently, we introduce synthetic data comprising reference images, modification texts, and target images to further enhance the compositional understanding capacity of the model and to narrow the gap between training and inference.

Specifically, we input the reference image $X_r$ and modification text $T_m$ into the model from Stage I, denoted as $\mathcal{F}_{stageI}$, using the template $P_2$ \textit{“A photo of \$ that $T_m$”} to encode the composed text feature $\mathbf{z_g} \in \mathbb{R}^{d}$. Simultaneously, we extract the target image features $\mathbf{t_g} \in \mathbb{R}^d$: 
\begin{gather}
    \mathbf{z_{cg}} = \mathcal{F}_{stageI}(X_r, T_m|P_2), \\
    \mathbf{t_g} = \mathcal{F}_{img}(X_t)
\end{gather}

Additionally, we also generate the mapping text features $\mathbf{z_g}$ using the template $P_1$ \textit{“A photo of \$”}, following a process that described in Section~\ref{sec:mapping}. Subsequently, we employ LLM to generate the caption $C_r$ corresponding reference image and extract the features $\mathbf{c_g}$:
\begin{gather}
    \mathbf{z_{mg}} = \mathcal{F}_{stageI}(X_r|P_1), \label{eq:map} \\
    \mathbf{c_g} = \mathcal{F}_{txt}(C_r)
\end{gather}
where $\mathcal{F}_{txt}$ is the original text encoder within the VLMs. Unless otherwise specified, “text encoder” refers to the text encoder with the VA that trained in Stage I.

\textbf{Composing Adapter.} 
Since the text encoder in VLMs is not explicitly trained to encode complex compositional semantics, we introduce lightweight composing adapters into the text encoder. These adapters are designed to learn how to effectively combine the pseudo-word token with real words, while keeping the backbone network frozen.

Specifically, for the $i$-th layer output representations $\mathbf{Z}^i \in \mathbb{R}^{l \times d}$ from the text encoder, we apply a down-projection followed by an up-projection through two MLPs,, defined as:
\begin{gather}
    \mathbf{\bar{Z}}^i = \psi_u(\psi_d(\mathbf{Z}^i))
\end{gather}
where $\psi_d$ and $\psi_u$ denote the down- and up-projection layers, respectively.  The output $\mathbf{\bar{Z}}^i$ is then added back to the original representation $\mathbf{Z}^i$ via a residual connection, and passed to the next layer of the text encoder.

\textbf{Optimization.} 
To facilitate the alignment between compositional text features and image features, we minimize the symmetric contrastive loss between the target image features $t_g$ and the composed text features $h_g$ , defined as:
\begin{gather}
     \mathcal{L}_{comp} = \mathcal{L}_{contra}(\mathbf{h_g}, \mathbf{t_g}) + \mathcal{L}_{contra}(\mathbf{t_g},\mathbf{h_g})
\end{gather}

In addition, we retain the objectives from Stage I to further enhance the mapping learning between the image and its pseudo-word representation, using the mapped text feature $\mathbf{z_g}$ and the corresponding caption feature $\mathbf{c_g}$. The overall training objective for Stage II is thus formulated as:
\begin{gather}
    \mathcal{L}_{\texttt{stage-II}} = \mathcal{L}_{comp} + \mathcal{L}_{map} + \alpha \mathcal{L}_{sta}
\end{gather}

\textbf{Hard Negative learning.} 
Given that the synthetic data is inherently limited and often noisy, we further introduce a hard negative mining strategy to enhance the model's discriminative ability. This approach aims to generate challenging yet diverse negative instances for objective $\mathcal{L}_{comp}$ through feature interpolation.
% Specifically, we start by selecting a set of top $k$ images, denoted as \mathbf{T} = $\{\mathbf{t}_1, \mathbf{t}_2, \dots, \mathbf{t}_k\}$ by calculating the similarity between the composed text features $\mathbf{z_{cg}}$ and the candidate target image features ${\mathbf{t_g}}$ from a mini-batch. Then, we calculate the similarity score between ground-truth target images and these top-$k$ candidate features as the mixing weights.
Specifically, we start by selecting a set of top-$k$ images, denoted as $\mathbf{T} = \{\mathbf{t}_1, \mathbf{t}_2, \dots, \mathbf{t}_k\}$, by calculating the similarity between the composed text features $\mathbf{z}_{\mathrm{cg}}$ and the candidate target image features $\mathbf{t}_{\mathrm{g}}$ from a mini-batch. Then, we calculate the similarity score between the ground-truth target images and these top-$k$ candidate features as the mixing weights.
Finally, $k$ novel hard negatives are generated by interpolating the features of the top $k$ target images with those of the ground-truth target images ${\mathbf{t_{gt}}}$, guided by the calculated weights. This process can be formulated as:

\vspace{-7mm}
\begin{gather}
    \mathbf{\bar{t}}_i=\beta_i \cdot \mathbf{t_{gt}} + (1-\beta_i) \cdot \mathbf{t}_i \\
    \beta = \texttt{Norm}(\mathbf{t_{gt}}^T\cdot \mathbf{T})
\end{gather}
where $\texttt{Norm}(u) =\frac{u-\min (u)}{\max (u)-\min (u)}$. This approach ensures that the resulting hard negatives strike a balance between being neither too similar nor too dissimilar to the query, thereby enhancing both the number and the complexity of negative samples.

\section{Experiments}

\subsection{Datasets and Metrics}
\noindent \textbf{Training:}
In Stage I, we train our method on the Conceptual Captions Three Million (CC3M) dataset \cite{sharma2018conceptual}, containing diverse image-caption pairs with no overlap with the evaluation datasets.
In Stage II, we leverage the synthetic triplet dataset generated by TransAgg \cite{liu2023zero}, which combines Laion-CIR-Template and LaionCIR-LLM, totaling around 32,000 triplets.

\noindent\textbf{Evaluation:}
We evaluate the performance on two standard benchmark datasets in ZS-CIR: FashionIQ~\cite{wu2021fashion}, CIRR~\cite{liu2021image} and CIRCO~\cite{baldrati2023zero}. 
\begin{itemize}
\item \textit{FishionIQ:} The dataset includes fashion items from three categories: Dresses, Shirts, and TopTees. Following prior studies \cite{saito2023pic2word, tang2024context}, we evaluated our model on the validation set, as the test set is not publicly available.
The validation set comprises a total of 6,000 triplets across the three categories. 
\item \textit{CIRR:} This dataset consists of approximately 21K real-life, open-domain images sourced from the NLVR2 dataset~\cite{suhr2018corpus}. We evaluated our model on the CIRR test set containing 4.1K testing triplets.
\item \textit{CIRCO:} 
This is an open-domain dataset recently developed from the COCO dataset to address the false negative issue. 
Unlike the aforementioned datasets, each sample in CIRCO consists of a reference image, a modification text, and multiple target images. We evaluated our model on the test set, which contains 800 triplets.
\end{itemize}

% Please add the following required packages to your document preamble:
% \usepackage{multirow}
\begin{table*}[tbp]
\caption{
Performance comparison on FashionIQ. “$\ddagger$” indicates methods where the model’s parameters are fully fine-tuned.
}
\centering 
\label{tab:fashion-iq}
\scalebox{1.0}{
\begin{tabular}{llcclcclccclll}
\toprule
 & \multirow{2}{*}{\textbf{Method}}   &\multirow{2}{*}{\textbf{Synthetic data}} & \multicolumn{2}{c}{\textbf{Dress}}       &  & \multicolumn{2}{c}{\textbf{Shirts}}      &  & \multicolumn{2}{c}{\textbf{TopTee}}      &  & \multicolumn{2}{c}{\textbf{Avg}}                    \\ \cline{4-5} \cline{7-8} \cline{10-11} \cline{13-14} 
 & &                        & R@10           & R@50           &  & R@10           & R@50           &  & R@10           & R@50           &  & \multicolumn{1}{c}{R@10}  & R@50           \\ \hline
 & Image-only      &{-}        & 4.86           & 12.99          &  & 11.04          & 20.22          &  & 8.67           & 16.52          &  & 8.19                      & 16.58          \\
 & Text-only    &{-}           & 14.33          & 32.57          &  & 20.46          & 33.61          &  & 21.72          & 39.32          &  & 18.84                     & 35.17          \\
 & Image + Text    &{-}        & 16.81          & 36.14          &  & 21.10          & 34.49          &  & 23.97          & 49.42          &  & 20.62                     & 36.69          \\
 & Captioning     &{-}         & 7.98           & 21.76          &  & 21.49          & 36.16          &  & 18.77          & 43.17          &  & 16.08                     & 30.70          \\
 & Pic2Word   \cite{saito2023pic2word}      &{-}         & 20.00          & 40.20          &  & 26.20          & 43.60          &  & 27.90          & 47.40          &  & 24.70  & 43.70          \\
 % & SEARLE-XL-OTI           & 21.57          & 44.47          &  & 30.37          & 47.49          &  & 30.90          & 51.76          &  & 27.61                    & 47.90          \\
 & SEARLE-XL    \cite{baldrati2023zero}      &{-}      & 20.48          & 43.13          &  & 26.89          & 45.58          &  & 29.32          & 49.97          &  & 25.56                    & 46.23          \\
 & KEDs    \cite{suo2024knowledge}    &{-}             & 21.70          & 43.80          &  & 28.90          & 48.00          &  & 29.90          & 51.90          &  & 26.80                     & 47.90          \\
 & LinCIR   \cite{gu2024language}    &{-}            & 20.92          & 42.44          &  & 29.10          & 46.81          &  & 28.81          & 50.18          &  & 26.28                     & 46.49          \\
 & ContextI2W    \cite{tang2024context}   &{-}         & 23.10          & 45.30          &  & 29.70          & 48.60          &  & 30.60          & 52.90          &  & 27.80                     & 48.90          \\
\rowcolor{mygray} & TSCIR  &{-}                 & \textbf{24.14} & \textbf{46.80} & & \textbf{31.01} & \textbf{50.05} & & \textbf{32.94} & \textbf{54.26} & & \textbf{29.37} & \textbf{50.37}      \\
 \hline
 & TransAgg \cite{liu2023zero}       &{32K}         & -              & -              &  & -              & -              &  & -              & -              &  & 28.57                     & 48.29          \\
 & TransAgg$\ddagger$ \cite{liu2023zero}    &{32K}        & -              & -              &  & -              & -              &  & -              & -              &  & 30.61                     & 50.38          \\
 & HyCIR   \cite{jiang2024hycir}    &{560k}              & 19.98                     & 40.80                     &  & 27.62                     & 44.94                     &  & 28.14                     & 47.64                     &  & 25.25                    & 44.46          \\
 & RTD   \cite{byun2024reducing}      &{2.5M}            &{23.50} &{46.65} &  & {27.97} &{46.96} &  & {31.31} & {53.09} &  & 27.59                    & 48.90          \\
& FTI4CIR   \cite{lin2024fine}     &{3M}           & {24.39} & {47.84} &  & {31.35} & {50.59} &  & {32.43} & {54.21} &  & 29.39                    & 50.88          \\
& CompoDiff \cite{gu2024compodiff}  &{18.8M}         & -              & -              &  & -              & -              &  & -              & -              &  & \textbf{36.02 }                    & 48.64        \\

 % &                         &                &                &  &                &                &  &                &                &  &                           &                \\
\rowcolor{mygray} & {TSCIR}          &{32K}          & \textbf{27.22} & \textbf{50.87} &  & \textbf{33.71} & \textbf{53.43} &  & \textbf{34.73} & \textbf{57.22} &  & {31.88}            & \textbf{53.84} \\
\bottomrule
\end{tabular}
}
\end{table*}

% ==========================================
% CIRR
% ==========================================

\begin{table}[tb]
\caption{Performance comparison on CIRR.}
\label{tab:cirr}
\centering 
\scalebox{0.8}{
\begin{tabular}{lcllll}
\toprule
\multirow{2}{*}{\textbf{Method}} &\multirow{2}{*}{\textbf{Synthetic data}}         & \multicolumn{4}{c}{\textbf{CIRR}}                                                          \\ \cline{3-6}  
              && R@1                       & R@5   & R@10                  & R@50                  \\ \hline
Image-only  &{-}   & 7.33                      & 23.01 & 33.25                 & 56.24                 \\
Text-only  &{-}    & 20.92                     & 43.98 & 55.42                 & 76.77                 \\
Image+Text &{-}    & \multicolumn{1}{l}{12.34} & 36.22 & 50.27                 & 78.15                 \\
Captioning &{-}    & \multicolumn{1}{l}{16.60} & 40.00 & 52.94                 & 79.33                 \\
Pic2Word      \cite{saito2023pic2word}    &{-}   & 23.90                     & 51.70 & 65.30                 & 87.80                 \\
SEARLE-XL     \cite{baldrati2023zero}  &{-}   & 24.24                     & 52.48 & 66.29                 & 88.84                 \\
KEDs    \cite{suo2024knowledge}     &{-}     & 26.40                     & 54.80 & 67.20                 & 89.20                 \\
LinCIR     \cite{gu2024language}     &{-}  & 25.04                     & 53.25 & 66.68                 & \multicolumn{1}{c}{-} \\
ContextI2W  \cite{tang2024context}   &{-}    & 25.60                     & 55.10 & 68.50                 & 89.80                 \\
\rowcolor{mygray}  TSCIR &{-}    &\textbf{26.10} &\textbf{55.15} &\textbf{68.66} & \textbf{90.06} \\
\hline
TransAgg  \cite{liu2023zero}  &{32k}   &{25.04} & 53.98 & \multicolumn{1}{c}{-} & \multicolumn{1}{c}{-} \\
TransAgg $\ddagger$ \cite{liu2023zero} &{32k} & {27.90} & 58.27 & \multicolumn{1}{c}{-} & \multicolumn{1}{c}{-} \\
HyCIR    \cite{jiang2024hycir}  &{560k}      & {25.08}          & 53.49          & 67.03                 & 89.85                 \\
RTD  \cite{byun2024reducing} &{2.5M}   & {27.86}          & {56.24}          &68.48                 & \multicolumn{1}{c}{-}                \\
FTI4CIR  \cite{lin2024fine}   &{3M}    & \multicolumn{1}{l}{25.90}          & 55.61                     & 67.66                     & 89.66    \\
CompoDiff \cite{gu2024compodiff } &{18.8M}  & {18.24}          & {-}          &-                 & \multicolumn{1}{c}{-} \\
\rowcolor{mygray}  TSCIR   &{32k}        & \textbf{29.16} & \textbf{59.33} & \textbf{71.88}        & \textbf{91.52}   \\    
\bottomrule
\end{tabular}
}
\end{table}

% Please add the following required packages to your document preamble:
% \usepackage{multirow}
\begin{table}[tb]
\caption{Performance comparison on CIRCO.}
\label{tab:circo}
\centering 
\scalebox{0.8}{
\begin{tabular}{lccccc}
\toprule
\multirow{2}{*}{\textbf{Method}} &\multirow{2}{*}{\textbf{Synthetic daa}} & \multicolumn{4}{c}{\textbf{CIRCO}}                                         \\ \cline{3-6} 
                       & & M@5            & M@10           & M@25           & M@50           \\ \hline
Image-only          &{-}    & 1.80           & 2.44           & 3.05           & 3.46           \\
Text-only      &{-}         & 3.01           & 3.18           & 3.68           & 3.93           \\
Image+Text    &{-}          & 4.32           & 5.24           & 6.49           & 7.07           \\
Captioning    &{-}          & 8.33           & 8.98           & 10.17          & 10.75          \\
Pic2Word   \cite{saito2023pic2word}   &{-}            & 8.72           & 9.51           & 10.46          & 11.29          \\
SEARLE-XL   \cite{baldrati2023zero}    &{-}          & 11.68          & 12.73          & 14.33          & 15.12          \\
LinCIR    \cite{gu2024language}    &{-}           & 12.59          & 13.58          & 15.00          & 15.85          \\
\rowcolor{mygray} TSCIR       &{-}            & \textbf{14.79}           & \textbf{15.15}          & \textbf{16.92}          & \textbf{19.00}          \\
\hline
HyCIR    \cite{jiang2024hycir}     &{560k}            & 14.12          & 15.02          & 16.72          & 17.56          \\
RTD   \cite{byun2024reducing}   &{2.5M}           & {9.13} & {9.63} & {10.68} & {11.27} \\
FTI4CIR    \cite{lin2024fine}    &{3M}           & 15.05          & 16.32          & 18.06          & 19.05          \\
CompoDiff \cite{gu2024compodiff} &{18.8M}  &12.31 &13.51 &15.67 &- \\
\rowcolor{mygray} TSCIR   &{32k}    & \textbf{18.37}          & \textbf{19.55}          & \textbf{21.64}          & \textbf{22.71}         \\
\bottomrule
\end{tabular}
}
\end{table}

\noindent\textbf{Metrics:}
Evaluation of the model primarily employs the Rank-K metric, which measures the probability of finding at least one target image within the top-K candidates based on a composed query. Specifically for CIRCO, the mean Average Precision (mAP) is the main criterion. Higher values in Rank-K and mAP indicate better performance.

\subsection{Implementation Details}
We use the visual and textual encoders of the CLIP ViT-L/14 \cite{radford2021learning} as our backbone. For optimization, we apply the AdamW optimizer with a learning rate of $10^{-4}$, a weight decay of 0.1, and a batch size of 128 for both stages. The temperature coefficient $\tau$ is set to 0.05 for Stage I and 0.07 for Stage II. The loss weight parameter $\alpha$ is set to 0.2 for both stages, and the adapter dimensionality is set to 128. 
All experiments were conducted on NVIDIA A100 GPUs.

\subsection{Comparison with State-of-the-Art Methods}
In Tables \ref{tab:fashion-iq}, \ref{tab:cirr}, and \ref{tab:circo}, we report the results on three benchmarks: Fashion-IQ, CIRR, and CIRCO. Among the state-of-the-art methods, we categorize them into two groups based on whether they utilize synthetic data generated by LLMs or diffusion models.
First, compared with methods that do not rely on synthetic data, our approach (i.e., Stage I) demonstrates significant improvements, highlighting the effectiveness of our mapping learning stage in capturing richer visual semantics and achieving better alignment with real-word tokens. Moreover, our method achieves competitive performance compared to approaches that rely on synthetic data.

On the other hand, we observe that introducing the second stage with synthetic data leads to a significant performance improvement. 
Compared with other methods that utilize synthetic data during training, our approach consistently achieves superior results across multiple benchmarks.
In particular, using the same synthetic dataset as TransAgg, our method outperforms it by a large margin, e.g., with a $11.59\%$ and $11.49\%$ improvements in terms of Avg on Fashion-IQ. Even compared to the fully fine-tuned version of TransAgg, our approach still offers significant advantages.

In addition, we compare \textit{the number of trainable parameters} in our method with current open-source pseudo-word token-based approaches. Among them, Pic2Word (1.64M), LinCIR (14.16M), and KEDs (10.50M) rely on global image features to generate pseudo-word tokens, which may limit their capacity to capture fine-grained visual semantics. 
In contrast, ContextI2W (68.47M) and FTI4CIR (76.61M) incorporate token learner modules to capture more comprehensive visual information.
Our method requires 21.27M (StageI: 20.09M + StageII: 1.18M) trainable parameters, striking a better balance between model complexity and retrieval performance.

\begin{table}[tbp]
\caption{Ablation study on the two-stage training framework.}
\label{ab:stage}
\centering 
\scalebox{1.0}{
\begin{tabular}{lllllll}
\toprule
\multirow{2}{*}{\textbf{Method}} & \multicolumn{2}{c}{\textbf{Fashion-IQ}} &  & \multicolumn{3}{c}{\textbf{CIRR}} \\ \cline{2-3} \cline{5-7} 
                         & R@10            & R@50           &  & R@1     & R@5     & R@10    \\ \hline
Baseline                &  24.80              & 44.32              &  & 24.01       & 51.53       & 65.12       \\
Stage I                   & 28.96          & 50.41         &  & 26.10  & 55.15  & 68.66  \\
Stage II   & 26.92          & 47.46         &  &25.22  &53.07   & 67.19\\ 
\rowcolor{mygray} Stage I+II                 &\textbf{31.88}          & \textbf{53.84}         &  & \textbf{29.16}       & \textbf{59.33}       &  \textbf{71.88}      \\

\bottomrule
\end{tabular}
}
\end{table}

\subsection{Ablation Studies}
In this section, we provide detailed ablation studies on Fashion-IQ and CIRR to verify the effectiveness of each part of our design. Moreover, we use our reproduced version of Pic2Word, which maps global features to pseudo-word tokens, as the Baseline method.

\noindent\textbf{Effectiveness of two-stage training.}
To evaluate the effectiveness of our proposed two-stage training framework, we conduct an ablation study by training each stage independently.
We observed inferior performance when solely training on Stage II, underscoring the crucial role of Stage I in establishing a robust image-to-word mapping to support subsequent composing learning in Stage II effectively.
While training Stage I alone achieves competitive performance, the incorporation of compositional learning in Stage II leads to significant performance enhancements. 
This demonstrates the complementary relationship between the two stages, i.e., first establishing a robust image-to-word mapping and then improving compositional understanding, and validates the rationality and effectiveness of the two-stage training framework.

% 请确保在导言区添加：
% \usepackage[table]{xcolor}

\begin{table}[tbp]
\caption{Ablation study on the proposed components. Each row in the table represents an incremental addition of components to the Baseline.
``VSI" indicates visual semantic injection module, ``${\mathcal{L}_{Stage I}}$" indicates $\mathcal{L}_{map}$ and $\mathcal{L}_{sta}$ in Equation \ref{eq:stage2_loss}, ``CA" indicates composing adapter, and ``HNL" indicates hard negative learning.}
\label{tab:component}
\centering 
\scalebox{1.0}{
\begin{tabular}{llllllll}
\toprule
\multirow{2}{*}{\textbf{Stage}}   & \multirow{2}{*}{\textbf{Method}} & \multicolumn{2}{c}{\textbf{Fashion-IQ}} &  & \multicolumn{3}{c}{\textbf{CIRR}} \\ \cline{3-4} \cline{6-8} 
                         &                         & R@10            & R@50           &  & R@1     & R@5     & R@10    \\ \hline
                         & Baseline                &  24.80              & 44.32              &  & 24.01       & 51.53       & 65.12       \\
\multirow{3}{*}{Stage I}  & +VSI                     & 27.12          & 48.65         &  & 25.30  & 53.98  &67.37  \\
                         % & +SA              & 25.93          & 46.48         &  &24.98        & 52.86       & 67.07       \\
                         & +$\mathcal{L}_{sta}$        & 28.96          & 50.41         &  & 26.10  & 55.15  & 68.66        \\ \hline
\multirow{2}{*}{Stage II} & +CA                    & 29.57          & 51.90         &  & 27.01       & 57.88       &  69.63     \\
& +$\mathcal{L}_{Stage I} $                & 30.09          & 53.00         &  & 28.57       & 58.93       &  70.01     \\
                          & \cellcolor{mygray}+HNL                  & \cellcolor{mygray}\textbf{31.21} & \cellcolor{mygray}\textbf{53.84} & \cellcolor{mygray} & \cellcolor{mygray}\textbf{29.16} & \cellcolor{mygray}\textbf{59.33} &  \cellcolor{mygray}\textbf{71.88}
     \\
\bottomrule
\end{tabular}
}
\end{table}

\noindent\textbf{Effectiveness of the proposed components.}
In Table \ref{tab:component}, we present an ablation study to validate the effectiveness of the proposed components. 
We observe consistent performance improvements with the incremental addition of components. Notably, introducing the VSI results in a significant performance boost, clearly demonstrating that injecting visual semantics into the text encoding process helps the model to capture more comprehensive image information.
Adding the soft text alignment objective further improves performance, highlighting the importance of alignment in the text embedding space.
In the second stage, incorporating a small amount of synthetic data along with the CA yields additional gains.
Moreover, applying the same objectives from Stage I to strengthen the mapping learning, along with the introduction of hard negative learning, leads to further notable improvements.

\begin{table}[tb!]
\caption{Ablation study on the injection position of visual semantics.}
\label{ab:token}
\centering 
\scalebox{1.0}{
\begin{tabular}{lllllll}
\toprule
\multirow{2}{*}{\textbf{Method}} & \multicolumn{2}{c}{\textbf{Fashion-IQ}} &  & \multicolumn{3}{c}{\textbf{CIRR}} \\ \cline{2-3} \cline{5-7} 
                         & R@10            & R@50           &  & R@1     & R@5     & R@10    \\ \hline
$[CLS]$ token                &  24.80              & 47.82              &  & 23.82       & 48.76       & 63.22       \\
All token                   & 28.96          & 46.00         &  & 22.24  & 46.25  & 62.16  \\
\rowcolor{mygray} Pseudo-word token                  & \textbf{31.88}          & \textbf{53.84}         &  & \textbf{29.16}       & \textbf{59.33}       &  \textbf{71.88}     \\

\bottomrule
\end{tabular}
}
\end{table}

\begin{table}[tb!]
\caption{Ablation study on the layer for visual semantic injection.}
\label{tab:va-position}
\centering 
\scalebox{1.0}{
\begin{tabular}{lllllll}
\toprule
\multirow{2}{*}{\textbf{Layer}} & \multicolumn{2}{c}{\textbf{Fashion-IQ}} &  & \multicolumn{3}{c}{\textbf{CIRR}} \\ \cline{2-3} \cline{5-7} 
                         & R@10            & R@50           &  & R@1     & R@5     & R@10    \\ \hline
% \{1,4\}                   & 29.16          & 50.91         &  &        &        &        \\
% \{4,7\}   \\ 
\{2,6\}   &27.63          & 51.01         &  & 25.02       & 55.89       &  69.24  \\
\{5,8\}   &28.12          & 51.97         &  & 26.01       & 56.70       &  69.94  \\ 
% \{1,3,5\}\\
\{2,6,10\}  &30.92          & 53.09         &  & 28.72       & 58.83       &  71.08     \\
\rowcolor{mygray} \{5,8,11\}                  & \textbf{31.88}          & \textbf{53.84}         &  & \textbf{29.16}       &\textbf{59.33}  &  \textbf{71.88}     \\ 
\{9,10,11\}                  & 31.22          & 53.03         &  & 29.00       & 59.13       &  71.67     \\ 
\{2,5,8,11\}  & 30.62          & 52.00         &  & 27.03       & 57.73       &  70.36     \\

\bottomrule
\end{tabular}
}
\end{table}

\begin{table}[tb!]
\caption{Ablation study on the position of the composed adapter.}
\label{tab:ca-position}
\centering 
\scalebox{1.0}{
\begin{tabular}{lllllll}
\toprule
\multirow{2}{*}{\textbf{Layer}} & \multicolumn{2}{c}{\textbf{Fashion-IQ}} &  & \multicolumn{3}{c}{\textbf{CIRR}} \\ \cline{2-3} \cline{5-7} 
                         & R@10            & R@50           &  & R@1     & R@5     & R@10    \\ \hline 
\{4,8,12\}     &30.59 &52.06 &&27.01 &58.11 &70.84                  \\
\{3,6,9,12\}   &31.41  &53.16 &&28.80 &59.13 &71.47\\
% \{4,6,8,10\} &28.23 &48.59\\
\rowcolor{mygray} \{2,4,6,8,10,12\}       & \textbf{31.88}          & 53.84         &  & 29.16       & 59.33       &  71.88     \\ 
% \{1,3,5,7,9,11\} \\
\{1-12\}        & {31.83}          & \textbf{54.04}         &  & \textbf{29.59}      & \textbf{59.52}       & \textbf{72.39}     \\

\bottomrule
\end{tabular}
}
\end{table}

\noindent\textbf{Effect on the injection position of visual semantics.}
As shown in Table~\ref{ab:token}, injecting visual semantics into all words or the [CLS] token results in a noticeable performance drop. This is because the reference image semantics tend to dominate the representation, overwhelming the compositional semantics and impairing accurate query generation.
In contrast, injecting visual semantics specifically into the pseudo-word token enables precise aggregation of the reference image’s visual content, while still promoting effective interaction with the modification text.

\noindent\textbf{Effect on the position of visual semantic injection module (VSI).}
Table \ref{tab:va-position} presents an analysis of the impact of the position of the VSI module within the text encoder to determine the optimal insertion layers. The results show that placing the VSI in shallow layers degrades model performance, likely due to the inability of shallow features to effectively capture and align visual semantics. Conversely, consistently injecting visual semantics into the middle and deep layers enhances the learning of the pseudo-word token. Notably, placing the visual semantic injection module in layers  \{5, 8, 11\} yields the best results, so we adopt this configuration for our experiments.

\noindent\textbf{Effect on the position of composing adapter (CA).}
Table \ref{tab:ca-position} examines the impact of the CA’s position within the text encoder. The CA is primarily designed to fine-tune the text encoder’s parameters, enabling it to generate composed queries by effectively combining the pseudo-word token with the modified text. The results indicate that incorporating more CAs leads to improved performance. To strike a balance between performance and the number of CAs, we adopt the configuration at positions \{2, 4, 6, 8, 10\}.

\begin{table}[tb!]
\caption{Ablation study on different backbones.}
\label{ab:backbone}
\centering 
\scalebox{1.0}{
\begin{tabular}{lllllll}
\toprule
\multirow{2}{*}{\textbf{Backbone}} & \multicolumn{2}{c}{\textbf{Fashion-IQ}} &  & \multicolumn{3}{c}{\textbf{CIRR}} \\ \cline{2-3} \cline{5-7} 
                         & R@10            & R@50           &  & R@1     & R@5     & R@10    \\ \hline
ViT-B                & 25.62         & 47.41         &  & 28.17  & 58.17  & 71.01  \\
ViT-L              &  31.88             & 53.84              &  & 29.16       & 59.33      & 71.88       \\
ViT-H                  & 35.13         & 55.61         &  & 31.25       & 61.95      &  74.51     \\

\bottomrule
\end{tabular}
}
\end{table}

\noindent\textbf{Effect on the different backbones.}
We conduct an ablation study to investigate the impact of different backbones.
In Table~\ref{ab:backbone}, the results demonstrate that our method is compatible with a range of backbone architectures, and consistently benefits from stronger ones, achieving improved retrieval performance.

\begin{table}[tb!]
\caption{Ablation study on  different types of synthetic data.
For SynTriplets18M, we randomly sample approximately $0.5\%$ of the dataset for training. Since the dataset does not provide modification texts (i.e., instructions), we adopt the provided target captions as substitutes for modification descriptions. }
\label{ab:dataset}
\centering 
\scalebox{0.9}{
\begin{tabular}{lllllll}
\toprule
\multirow{2}{*}{\textbf{Synthetic Data}} & \multicolumn{2}{c}{\textbf{Fashion-IQ}} &  & \multicolumn{3}{c}{\textbf{CIRR}} \\ \cline{2-3} \cline{5-7} 
                         & R@10            & R@50           &  & R@1     & R@5     & R@10    \\ \hline
w/o                   & 28.96          & 50.41         &  & 26.10  & 55.15  & 68.66  \\
SynTriplets18M(87k)               & 30.44         & 51.92         &  & 27.69  & 56.97  & 69.71  \\
Laion-CIR(32k)            &  31.88              & 53.84              &  & 29.16       & 59.33      & 71.88       \\

\bottomrule
\end{tabular}
}
\end{table}

\noindent \textbf{Effect on the different types of synthetic data.}
Due to data availability constraints, we perform experiments on two open-source datasets, SynTriplets18M~\cite{gu2024compodiff} and Laion-CIR~\cite{liu2023zero}.
We conduct experiments on them to evaluate the performance and generalization ability of our method. 
Since SynTriplets18M does not provide modification texts (i.e., instructions), we use the provided target captions in the dataset as substitutes for modification texts. In our experiments, we randomly sample 87K triplets from SynTriplets18M rather than using the full dataset.
Notably, SynTriplets18M adopts diffusion models to synthesize the target images, whereas Laion-CIR generates them via a retrieval-based approach.
The overall quality of samples in SynTriplets18M is relatively lower compared to Laion-CIR.
Nevertheless, our method benefits from the inclusion of a small amount of synthetic data in both cases, achieving consistent performance improvements regardless of the data quality.

\noindent \textbf{Effect on the parameter $\alpha$.}
In Table~\ref{ab:alpha}, we explore the impact of the parameter $\alpha$ on performance. This parameter controls the weight of the soft text alignment loss, which helps the model combine the pseudo-word token with the modified text. When $\alpha$ is set to a larger value, performance decreases. We speculate that this may be due to the larger value over-constraining the model, causing it to overfit to the semantically limited text descriptions. When $\alpha = 0.2$, the model achieves better results, so we set it to 0.2 in our experiments.

\begin{table}[tb!]
\caption{Ablation study on parameter $\alpha$.}
\label{ab:alpha}
\centering 
\scalebox{1.0}{
\begin{tabular}{cllllll}
\toprule
\multirow{2}{*}{\textbf{Parameter $\alpha$}} & \multicolumn{2}{c}{\textbf{Fashion-IQ}} &  & \multicolumn{3}{c}{\textbf{CIRR}} \\ \cline{2-3} \cline{5-7} 
                         & R@10            & R@50           &  & R@1     & R@5     & R@10    \\ \hline
0.0                &  27.45              & 50.68              &  & 25.91       & 57.03       & 69.62       \\
\rowcolor{mygray} 0.2                   &\textbf{31.88}          & \textbf{53.84}         &  & \textbf{29.16}       & \textbf{59.33}       &  \textbf{71.88} \\
0.4   & 30.97         & 53.06         &  &28.65  &58.77   & 71.10\\ 
0.6                &{28.18}          & {52.30}         &  & {26.94}       & {57.08}       &  {70.53}      \\

\bottomrule
\end{tabular}
}
\end{table}

\begin{table}[tb!]
\caption{Ablation study on parameter $k$.}
\label{ab:k}
\centering 
\scalebox{1.0}{
\begin{tabular}{cllllll}
\toprule
\multirow{2}{*}{\textbf{Parameter $k$}} & \multicolumn{2}{c}{\textbf{Fashion-IQ}} &  & \multicolumn{3}{c}{\textbf{CIRR}} \\ \cline{2-3} \cline{5-7} 
                         & R@10            & R@50           &  & R@1     & R@5     & R@10    \\ \hline
0                & 30.09          & 53.00         &  & 28.57       & 58.93       &  70.01     \\
10                  & {30.67}          & {53.26}         &  & {28.99}       & {59.23}       &  {71.34}      \\ 
\rowcolor{mygray} 20   &\textbf{31.88}          & \textbf{53.84}         &  & \textbf{29.16}       & \textbf{59.33}       &  \textbf{71.88} \\
30                &{30.21}          & {53.30}         &  & {28.76}       & {59.20}       &  {71.52}      \\

\bottomrule
\end{tabular}
}
\end{table}

\begin{figure*}[tb!]
\centering\includegraphics[width=1.8\columnwidth]{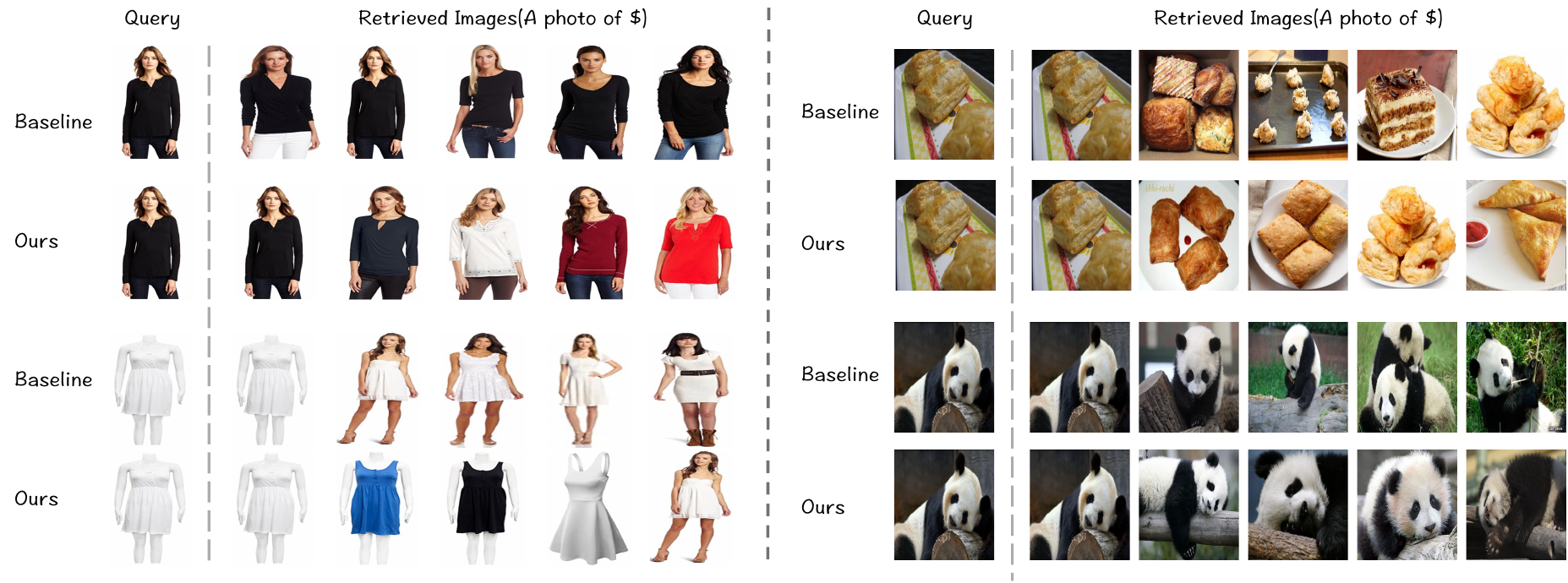}
% \vspace{-6mm}
\caption{Retrieved results of "A photo of \$" in Stage I on Fishion-IQ (left) and CIRR (right). ``\$" indicates the pseudo-word token generated by the query image. 
}\label{fig:retrieval-mapping}
% \vspace{-4mm}
\end{figure*}

\noindent \textbf{Effect on the parameter $k$.}
In Table~\ref{ab:k}, we investigate the impact of the parameter $k$, which determines the number of mixup candidates, on performance. Since the synthetic data is limited, we introduce the hard learning strategy to enhance the model’s discriminative ability. The results show that incorporating the hard learning strategy leads to additional performance gains. However, increasing the number of candidates did not result in further improvements. We speculate that selecting more top-k candidates reduces the similarity between the later candidates and the ground truth, preventing the mixup from generating hard negatives. In our experiments, we set $k=20$.

\begin{figure}[tb!]
\centering\includegraphics[width=1.0\columnwidth]{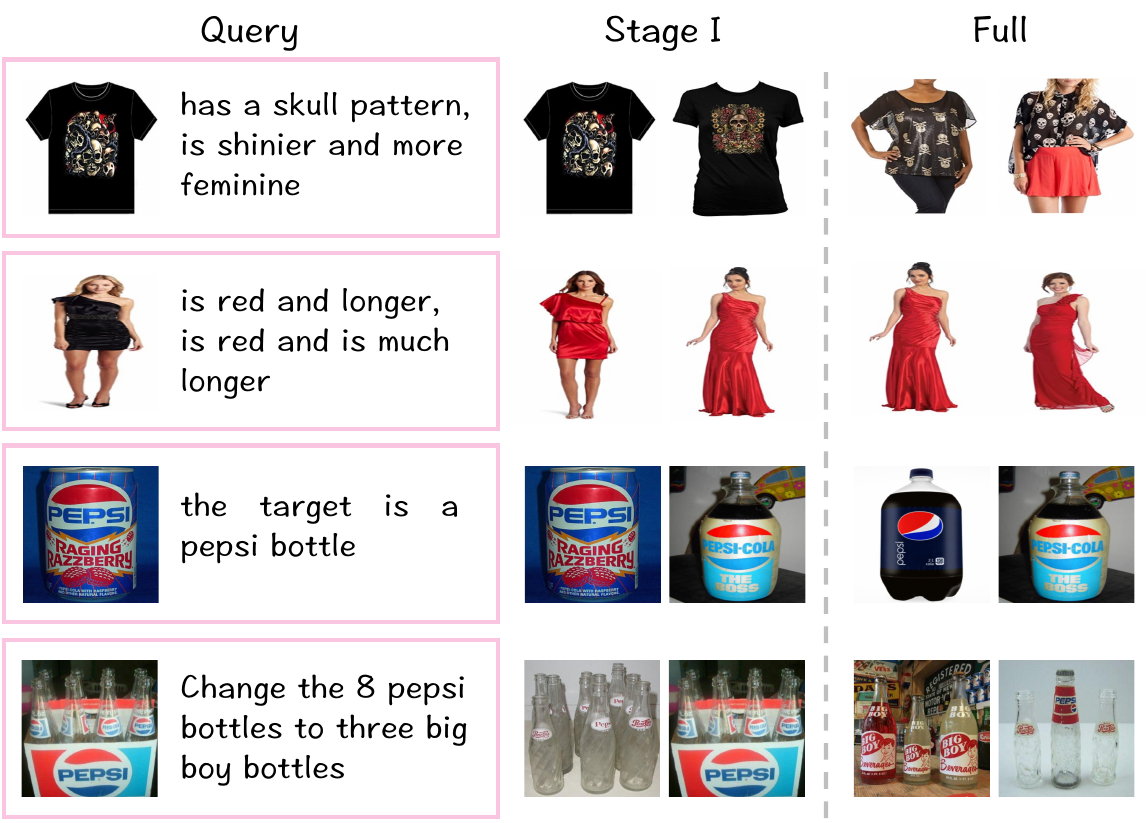}
\vspace{-6mm}
\caption{Illustration of composed image retrieval on Fashion-IQ and CIRR.
}\label{fig:retrieval-composed}
% \vspace{-4mm}
\end{figure}

\subsection{Qualitative Analysis}
\noindent \textbf{Retrieval results in Stage I.} 
To validate the effectiveness of mapping learning in Stage I, we present retrieval results for the query “A photo of \$.” The results show that our method accurately captures the overall image content (e.g., multiple subjects) while preserving rich, detailed information.
For example, in the first case on the left, our method identifies intricate details, such as the collar design, and retrieves visually consistent images with similar structural elements, like the blue dress in the second case. In contrast, the baseline only focuses on simpler attributes, such as color. 
We attribute this to the visual semantic injection module's ability to continuously inject visual semantics, enabling the model to capture both global content and fine-grained details.

\noindent \textbf{Composed image retrieval results.}
Figure~\ref{fig:retrieval-composed} presents the composed image retrieval results of our Full and Stage I models on Fashion-IQ.
In the first case, the Stage I model focuses on the skull pattern but overlooks modifications like “shinier” and “feminine,” highlighting its limited ability to generate compositional semantics by integrating information from the reference image and modification text. This limitation stems from its reliance on mapping images to pseudo-word tokens and the retrieval ability of the pre-trained vision-language model.
In contrast, our Full model effectively captures the semantics of the reference image and integrates them with the modification text for accurate target image retrieval.

\section{Conclusion}

In this work, we propose a novel two-stage framework, TSCIR, for zero-shot Composed Image Retrieval, transitioning the training process from mapping to composing.
In the first stage, we focus on visual semantic mapping by introducing a visual semantic injection module, which continuously injects visual information into the text encoding process. Additionally, we employ a soft text alignment objective to align pseudo-word tokens with real-word semantics.
In the second stage, we incorporate composed adapters and a hard negative learning strategy to enhance the model’s ability to combine pseudo-word tokens with modification texts, enabling the generation of accurate composed queries.
Moreover, our method is compatible with both high-quality and low-quality synthetic data, and can achieve additional performance gains using only a small amount of synthetic data.
Extensive experiments on three CIR benchmarks demonstrate the effectiveness and generalization of our approach.

% \section*{Acknowledgments}
% This should be a simple paragraph before the References to thank those individuals and institutions who have supported your work on this article.

% {\appendix[Proof of the Zonklar Equations]
% Use $\backslash${\tt{appendix}} if you have a single appendix:
% Do not use $\backslash${\tt{section}} anymore after $\backslash${\tt{appendix}}, only $\backslash${\tt{section*}}.
% If you have multiple appendixes use $\backslash${\tt{appendices}} then use $\backslash${\tt{section}} to start each appendix.
% You must declare a $\backslash${\tt{section}} before using any $\backslash${\tt{subsection}} or using $\backslash${\tt{label}} ($\backslash${\tt{appendices}} by itself
%  starts a section numbered zero.)}

%{\appendices
%\section*{Proof of the First Zonklar Equation}
%Appendix one text goes here.
% You can choose not to have a title for an appendix if you want by leaving the argument blank
%\section*{Proof of the Second Zonklar Equation}
%Appendix two text goes here.}

% \section{References Section}
% You can use a bibliography generated by BibTeX as a .bbl file.
%  BibTeX documentation can be easily obtained at:
%  http://mirror.ctan.org/biblio/bibtex/contrib/doc/
%  The IEEEtran BibTeX style support page is:
%  http://www.michaelshell.org/tex/ieeetran/bibtex/
 
 % argument is your BibTeX string definitions and bibliography database(s)
\bibliography{egbib}
%
% \section{Simple References}
% You can manually copy in the resultant .bbl file and set second argument of $\backslash${\tt{begin}} to the number of references
%  (used to reserve space for the reference number labels box).

% \begin{thebibliography}{1}
\bibliographystyle{IEEEtran}

% \bibitem{ref1}
% {\it{Mathematics Into Type}}. American Mathematical Society. [Online]. Available: https://www.ams.org/arc/styleguide/mit-2.pdf

% \bibitem{ref2}
% T. W. Chaundy, P. R. Barrett and C. Batey, {\it{The Printing of Mathematics}}. London, U.K., Oxford Univ. Press, 1954.

% \bibitem{ref3}
% F. Mittelbach and M. Goossens, {\it{The \LaTeX Companion}}, 2nd ed. Boston, MA, USA: Pearson, 2004.

% \bibitem{ref4}
% G. Gr\"atzer, {\it{More Math Into LaTeX}}, New York, NY, USA: Springer, 2007.

% \bibitem{ref5}M. Letourneau and J. W. Sharp, {\it{AMS-StyleGuide-online.pdf,}} American Mathematical Society, Providence, RI, USA, [Online]. Available: http://www.ams.org/arc/styleguide/index.html

% \bibitem{ref6}
% H. Sira-Ramirez, ``On the sliding mode control of nonlinear systems,'' \textit{Syst. Control Lett.}, vol. 19, pp. 303--312, 1992.

% \bibitem{ref7}
% A. Levant, ``Exact differentiation of signals with unbounded higher derivatives,''  in \textit{Proc. 45th IEEE Conf. Decis.
% Control}, San Diego, CA, USA, 2006, pp. 5585--5590. DOI: 10.1109/CDC.2006.377165.

% \bibitem{ref8}
% M. Fliess, C. Join, and H. Sira-Ramirez, ``Non-linear estimation is easy,'' \textit{Int. J. Model., Ident. Control}, vol. 4, no. 1, pp. 12--27, 2008.

% \bibitem{ref9}
% R. Ortega, A. Astolfi, G. Bastin, and H. Rodriguez, ``Stabilization of food-chain systems using a port-controlled Hamiltonian description,'' in \textit{Proc. Amer. Control Conf.}, Chicago, IL, USA,
% 2000, pp. 2245--2249.

% \end{thebibliography}

\newpage

% \section{Biography Section}
% If you have an EPS/PDF photo (graphicx package needed), extra braces are
%  needed around the contents of the optional argument to biography to prevent
%  the LaTeX parser from getting confused when it sees the complicated
%  $\backslash${\tt{includegraphics}} command within an optional argument. (You can create
%  your own custom macro containing the $\backslash${\tt{includegraphics}} command to make things
%  simpler here.)
 
% \vspace{11pt}

% \bf{If you include a photo:}\vspace{-33pt}
% \begin{IEEEbiography}[{\includegraphics[width=1in,height=1.25in,clip,keepaspectratio]{fig1}}]{Michael Shell}
% Use $\backslash${\tt{begin\{IEEEbiography\}}} and then for the 1st argument use $\backslash${\tt{includegraphics}} to declare and link the author photo.
% Use the author name as the 3rd argument followed by the biography text.
% \end{IEEEbiography}

% \vspace{11pt}

% \bf{If you will not include a photo:}\vspace{-33pt}
% \begin{IEEEbiographynophoto}{John Doe}
% Use $\backslash${\tt{begin\{IEEEbiographynophoto\}}} and the author name as the argument followed by the biography text.
% \end{IEEEbiographynophoto}

\vfill

\end{document}